\documentclass[conference]{IEEEtran}
\IEEEoverridecommandlockouts
\usepackage{cite}
\usepackage{amsmath,amssymb,amsfonts}
\usepackage{algorithmic}
\usepackage{url} 
\usepackage{graphicx}
\usepackage{textcomp}
\usepackage{xcolor}
\usepackage{booktabs}   
\usepackage{multirow}   
\usepackage{overpic}
\def\BibTeX{{\rm B\kern-.05em{\sc i\kern-.025em b}\kern-.08em
    T\kern-.1667em\lower.7ex\hbox{E}\kern-.125emX}}
\begin{document}

\title{Training a Conditioned Video Game Agent on a VLM Annotated Dataset}

\author{\IEEEauthorblockN{Katrin Schmid}
\IEEEauthorblockA{\textit{NVIDIA}, Australia \\
kschmid@nvidia.com}
\and
\IEEEauthorblockN{Iuri Frosio}
\IEEEauthorblockA{\textit{NVIDIA}, Italy \\
ifrosio@nvidia.com}
}


\maketitle

\begin{abstract}
Reinforcement Learning (RL) is a powerful but far from easy-to-use technique for policy learning. In the specific case of video games, access to the game engine is required to get rewards for training (e.g. to collect rewards from the environment). Furthermore, the proper identification and weighting of the rewards generally requires a difficult trial-and-error approach. Lastly, rewards are often sparse and understanding how they eventually affect the learned policy is a non-trivial exercise. To ease these issues we propose annotating a video game dataset with Vision Language Models (VLMs) instructed to extract human defined rewards. We show that offline RL can then be used to train a conditioned agent that responds accordingly to the desired returns and  we discuss the difficulties and limitations that emerged in our early experiments.
\end{abstract}


\section{Introduction}

\begin{figure}
    \centering
\includegraphics[width=0.7\linewidth, trim=1.5cm 2.5cm 1.5cm 3cm, clip]{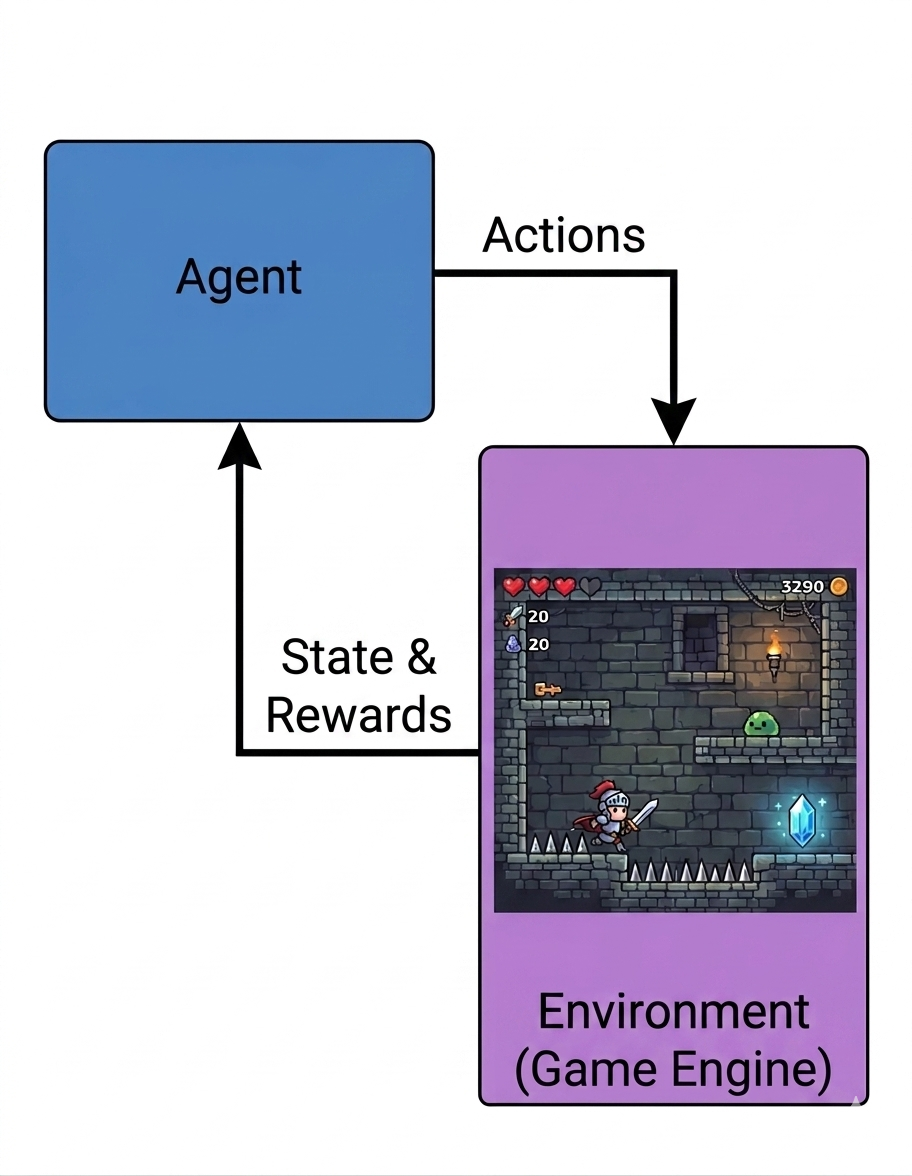}
    \caption{Typical RL setup in video games.}
    \label{fig:RL}
\end{figure}

In many contexts, including video games, RL is a viable solution to learn policies aimed at solving a target task.
In the typical setup an agent interacts with a real (or artificial, in case of video games) environment to collect a set of rewards as a consequence of its action (Fig.~\ref{fig:RL}); typical rewards for video games may be the score increasing after shooting an enemy or a loss of life at the end of a race after crashing a car against a wall.
The practical issues for the adoption of RL in this context are numerous though. First of all, access to the game engine is needed to collect the rewards; alternative solutions such as memory sniffing or game re-implementation~\cite{Dalton2019AcceleratingRL} are possible, but complex. The rewards can then be sparse (e.g., enemy killing may be a rare event in games like Counter-Strike~\cite{durst2024csgo}), making RL hard because of the credit assignment problem~\cite{Sutton1998}. 
When multiple partial rewards are available, the RL user has to decide how to combine them to define an expected return that has to be maximized; for example, what is the value of capturing a knight or a pawn in chess? Over-weighting pawn may lead to a policy that favors immediate, small rewards while remaining unable to win the final game. Approaches like inverse reinforcement learning~\cite{ng2000algorithms} have emerged to solve this issue, but even when effective, they do not guarantee full explainability of the reward system.

Here we leverage AI to try easing these issues. Our idea is based on the observation that rewards may be difficult to extract from the environment and translate numerically, but easy to describe in words by experts. Therefore we leverage Vision Language Models (VLMs) as in~\cite{2026annotation} to annotate an offline dataset with such rewards. Our contribution here is:
\begin{itemize}
    \item to demonstrate on a toy problem that such annotated dataset can be used for offline RL; the trained agent is capable of estimating the same set of rewards in real time: it gains  a deep understanding of the target environment and can be used as cheap annotator;
    \item to show that a conditioned agent responds to the user requests according to the set of adopted rewards;
    \item to study the issues that emerged in our preliminary experiments, especially regarding the sparsity of the rewards and the difficulty of training.
\end{itemize}

\section{Related Work}

For the task of dataset annotation we refer to~\cite{2026annotation}.
The difficulty of training with sparse rewards is well described in the literature~\cite{Sutton1998}. Another research theme  related to our paper is the estimate of an implicit reward function starting from a set of demonstrations, that is classically tackled in the context of inverse RL~\cite{ng2000algorithms}. We highlight here that our approach is not to be seen as a pure alternative to inverse RL: we believe, on the other hand, that leveraging the pros of the two approaches has great potential for future research. Our work shares commonalities also with approaches based on implicit curiosity and inverse reinforcement learning, like GAIL~\cite{ho2016gail} and HER~\cite{andrychowicz2017her}.  Lastly, we want to mention training of conditioned models, as we share affinity with Conditional Behavioral Cloning~\cite{codevilla2018cbc}, Decision Transformer~\cite{NEURIPS2021_7f489f64} and Multi-Game Decision Transformers~\cite{lee2022multigame}.

\section{Method}

\begin{figure*}[!t]
 \begin{center}
    \includegraphics[width=0.15\linewidth]{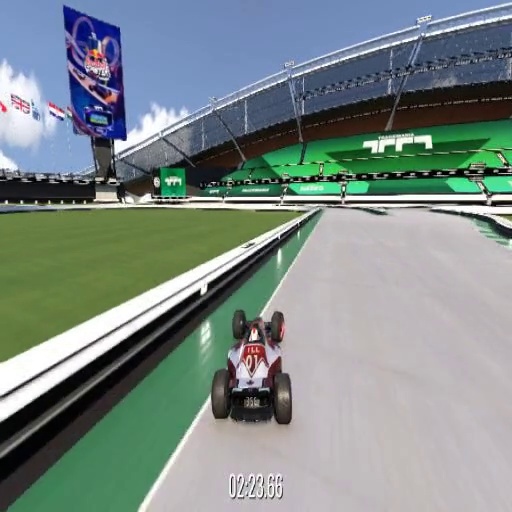}
    \includegraphics[width=0.15\linewidth]{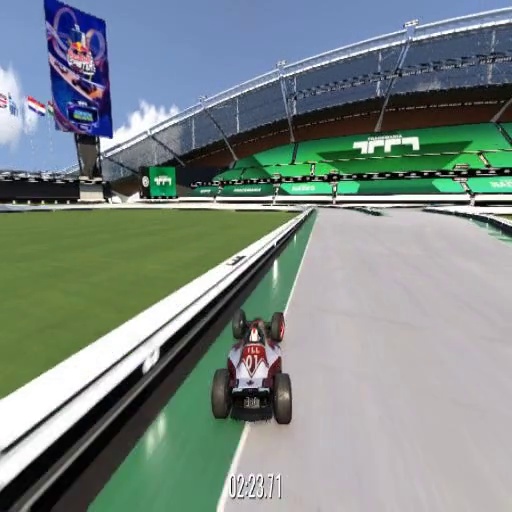}
    \includegraphics[width=0.15\linewidth]{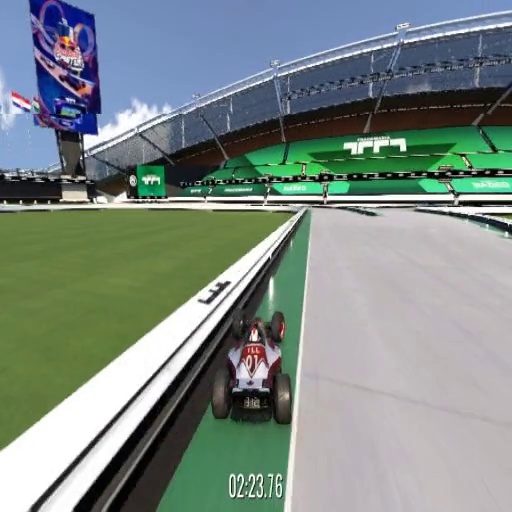}
    \includegraphics[width=0.15\linewidth]{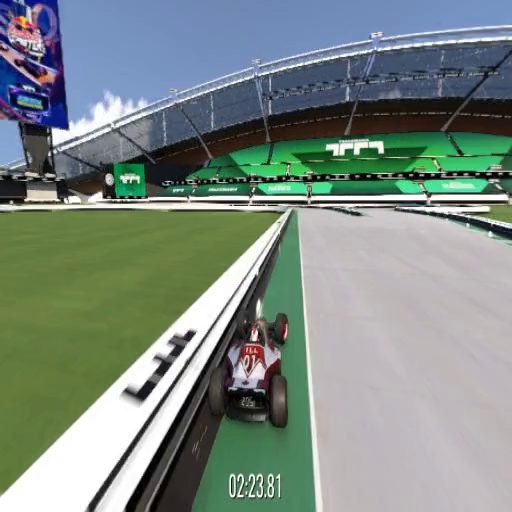}
    \includegraphics[width=0.15\linewidth]{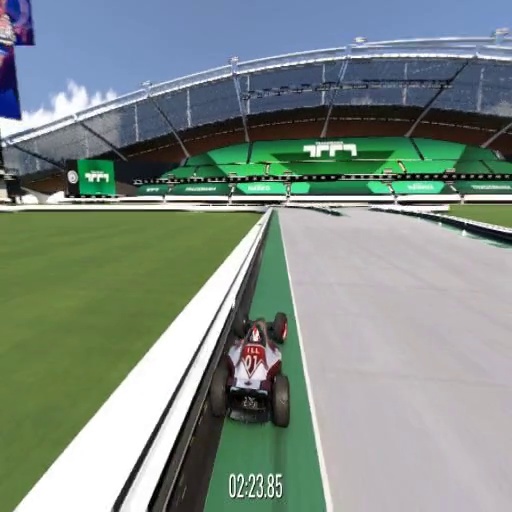}
    \includegraphics[width=0.15\linewidth]{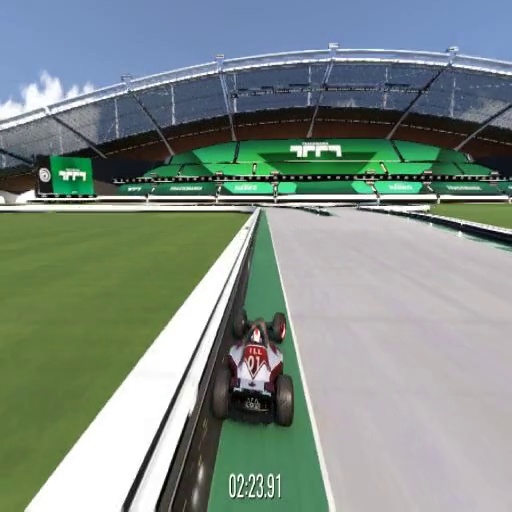}

    \end{center}
    \caption{A sequence from Trackmania used here for testing.}
    \label{fig:trackmania}
\end{figure*}

We performed experiments on Trackmania, a racing video game that allowed us to design a short track with clear visuals of the racing car and without distracting elements that would make training harder.
A typical sequence is shown in Fig.~\ref{fig:trackmania}.
As a first step we acquired a dataset of human game play consisting of Trackmania frames and key press sequences.
The dataset includes 25,000 frames at $512\times 512$\,px resolution, captured at $20$\,Hz.
It was later annotated with Qwen-3.5~\cite{qwen35omni} once every 6 frames (corresponding to 0.3 s intervals).
More specifically, we asked Qwen-3.5 to answer a set of three questions $\{Q_j\}_{j=1.. 3}$ for each 6 frame sequence; answers are probability in the 0-100\% range:

\begin{itemize}
    \item $Q_1$: is the red and white car in the middle of the road?
    \item $Q_2$: is the red and white car moving forward?
    \item $Q_3$: is the red and white car hitting the lateral barriers?    
\end{itemize}

The choice of these questions was based on several factors: first of all, their importance in the context of the gaming economy.
Moving forward or hitting a barrier can be easily associated with a positive or negative reward in  Trackmania.
Driving in the middle of the road is instead connected to the driving style and not necessarily to performance: when the car is in the middle of the road, it hardly hits the lateral barriers, but at the same time it cannot grab the curbs in the curves for maximum speed.
Minimizing or maximizing such reward changes in practice the driving style of the agent.
Each question $Q_j$ can therefore be associated to a reward $R_j$ that can be used for policy learning with RL and later for conditioning the agent behavior.
A second factor was human interpretability of the rewarfs.
The last one was the VLM's capability to extract such information from visual evidence.
The VLM output were finally interpolated along the temporal axis to create a dense set of rewards for training.

For each reward, we defined a discounted reward (also called \emph{return}) on a $1.5$\,s temporal window (corresponding to 30 frames), more formally defined as:
\begin{equation}
r_j(t) = \sum_{k=0}^{30}{\hat{R}_j(t+k)\cdot\gamma^k},
\label{eq:return}
\end{equation}
where we set $\gamma = 0.99$. The return $r_j(t)$ is associated with the car behavior in the next $1.5$\,s: for instance, if the car remains in the middle of the road, $r_1$ is maximum. If, on the other hand, the car hits a barrier in the next 1.5, $r_3 > 0$.
It is worthy noting that the return is computed on a limited interval of time and not summing the discounted rewards on the entire episode (as it is typical in RL): this is because we want to condition the agent behavior in a limited time window and not for the entire length of the episode.

\begin{figure}[h]
  \centering  \includegraphics[width=0.7\linewidth,trim=0cm 1cm 0cm 1.4cm, clip]{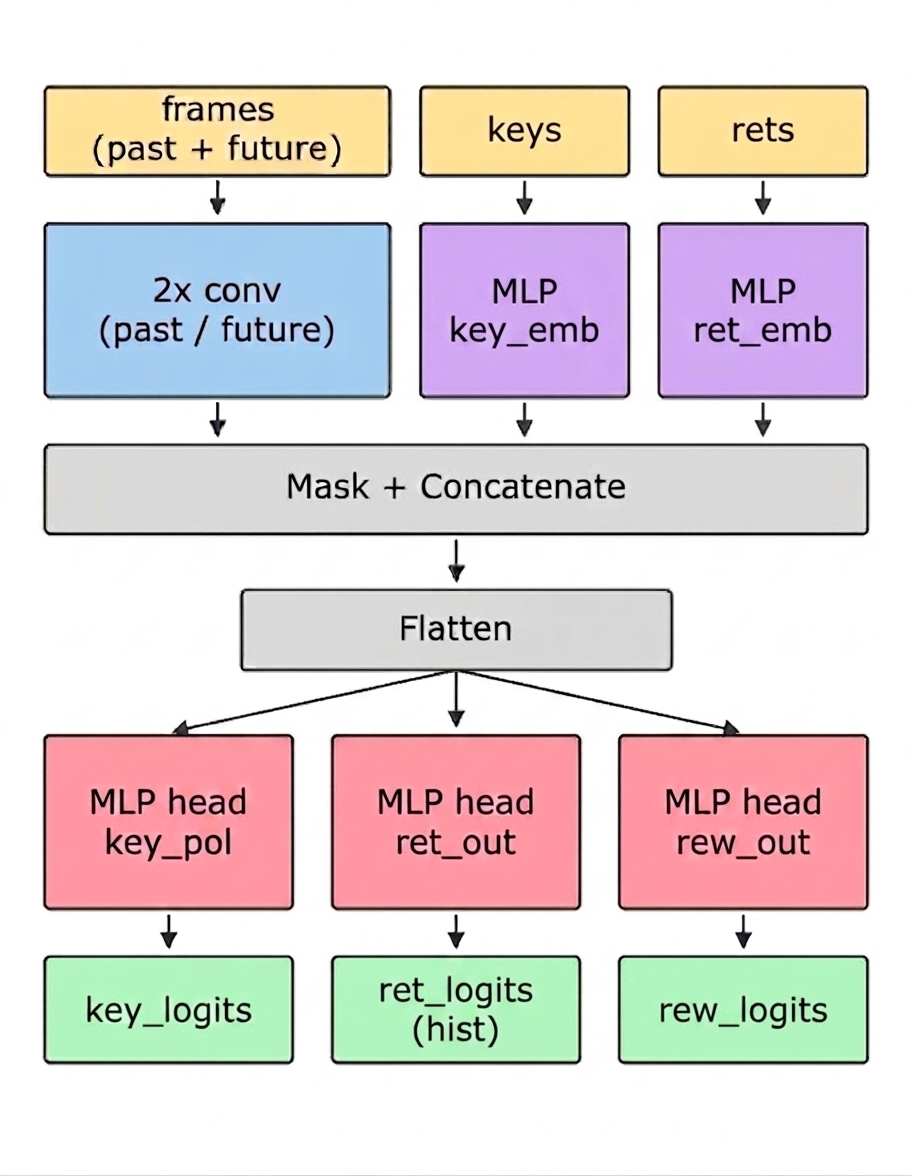}
  \caption{Architecture of the trained agent. When the return input is omitted, it performs unconditioned
  behavioral cloning.}
  \label{fig:rcnn}
\end{figure}

We provide returns as conditioning input to our agent, similarly to decision transformers\cite{NEURIPS2021_7f489f64, lee2022multigame}.
Inputs to our model are (Fig.~\ref{fig:rcnn}): the set of past 6 frames (resampled at $128\times 128$\,px resolution) and key presses, plus (for a conditioned agent only) the set of 3 desired returns. The frames go through a set of three convolution layers with leaky-ReLU activations, while the input keys and desired returns are processed with two sets of linear layers and leaky-ReLU activations. These are then flattened into a set of tokens with embedding size 16 and passed to 2 linear layers with leaky-ReLU activations to output the logits of the key presses in the next 2 frames; similar MLPs are used to predict the distribution of the next expected returns and rewards.
We train the agent on a loss including the binary cross-entropy on each future key press and on the desired reward and returns, plus a small regularization term to avoid collapse of key entropy.
Training is performed with batch size 128 and approximately 6000 training steps that we found sufficient (although further training could improve the quality of the conditioned agents). When the desired returns are not provided as input, the model is performing simple behavioral cloning (which we use as a baseline). Optimization is performed using RMSProp with learning rate 0.001.

\section{Results and Discussion}

\subsection{Ground truth rewards}

One of the reasons for choosing Trackmania as a testbed is that it can be coupled with Openplanet~\cite{openplanet} to access internal game information. We leveraged this feature by writing a plugin that exposes the 3D position and speed of the car on the track; this allows us to compute ground truth rewards $\{R_j\}_{j=1..3}$ associated with each question and thus evaluate numerically the performance of our conditioned agent.
It is important to highlight that we did not use this information for training in any way: ground truth rewards were used only for quantitative evaluation.
For computing them we proceeded as follows:
first, we captured the 3D position of the left and right side boundaries of the track and sampled them at uniform steps to get two lists of 1000 points, $\{{\bf{P}}^L_k\}_{k=1..1000}$ and $\{{\bf{P}}^R_k\}_{k=1..1000}$. When the car is in position ${\bf{P}}^C$, we find the closest ${\bf{P}}^L_k$ and ${\bf{P}}^R_k$.
We estimate the ground-truth probability that the car is in the middle of the road,  $R_1$, from the projection of $P_C$ onto the segment $\bf{P}^L_k–\bf{P}^R_k$. When the car is within the central 40\% of the segment, $R_1 = 1$; outside this band, $R_1$ decays linearly to 0 at the edges. 
The car is not moving forward ($R_2$) when either its gear is back, or the speed reported by OpenPlanet is below a threshold.
As for the car hitting the barrier ($R_3$), we assume that the car is against the wall when its position is less than a threshold from the closest ${\bf{P}}^L_k$ or ${\bf{P}}^R_k$ point.
The ground truth rewards are used to quantify the capability of our trained agent to output the correct rewards as well as to measure how much the agent respects the indications provided as conditioning input. 

\subsection{Evaluation}

We first evaluate the capability of our conditioned agent to act as a reward estimator. We compare each reward output by the agent to the ground truth reward computed as in the previous section, and average them over 100 runs over the test track. The right side of Table I shows the accuracy, precision, recall and $F1$ for the trained agent in two different conditioning states: when is asked to drive the car in the middle of the road ($r_1$ is maximum) and when, on the contrary, the first desired return is set to 0 (and thus the car is supposed to drive far from the middle of the road).
The same Table reports in the left columns the average value of the ground truth rewards: this indicates the percentage of time in our experiments when the car is driving in the middle of the road  ($R_1$), is moving forward ($R_2$) or hitting the barriers ($R_3$).

\begin{table*}
\centering
\caption{Average collected rewards $R_j$ and accuracy, precision, recall, and $F1$ on the estimated rewards for different agents.}
\label{tab:metrics}
\scriptsize
\begin{tabular}{c ccc cccc cccc cccc}
\toprule
\multirow{2}{*}{Conditioning} &
\multirow{2}{*}{Middle ($R_1$)} &
\multirow{2}{*}{Forward ($R_2$)} &
\multirow{2}{*}{Barrier($R_3$)} &
\multicolumn{4}{c}{Middle ($Q_1$)} &
\multicolumn{4}{c}{Forward ($Q_2$)} &
\multicolumn{4}{c}{Barrier ($Q_3$)} \\
\cmidrule(lr){5-8}\cmidrule(lr){9-12}\cmidrule(lr){13-16}
 &&&& Acc & Pre & Rec & $F1$
 & Acc & Pre & Rec & $F1$
 & Acc & Pre & Rec & $F1$\\
\midrule
Minimize Middle & 37\% & 87\% & 11\% & 75\% & 61\% & 87\% & 72\% & 93\% & 94\% & 99\% & 96\% & 86\% & 29\% & 23\% & 26\% \\
Maximize Middle & 43\% & 87\% & 9\% & 61\% & 52\% & 98\% & 68\% & 93\% & 93\% & 99\% & 96\% & 91\% & 38\% & 8\% & 13\%\\
None & 45\% & 87\% & 5\% & --- & --- & --- & --- & --- & --- & --- & --- & --- & --- & --- & --- \\
\bottomrule
\end{tabular}
\end{table*}

$R_2$ is predicted accurately ($F1>90\%$).
With the car moving forward $87\%$ of the time, $R_2$ is a strongly class-imbalanced target: high accuracy here is expected but not guaranteed.
The prediction for $R_1$ is slightly less accurate, but the metris demonstrate that the model clearly learns to predict the position of the car; further improvements likely require the use of larger training sets or a different agent architecture.
The estimate for $R_3$ is more problematic: from ground truth data we know that the car is hitting the barrier only 9\% or 11\% of the time, which makes training data for $R_3$ unbalanced (although $R_2$ is unbalanced as well).
An additional difficulty for $R_3$ may arise from the fact that a barrier hit may take as little as a few frames, while our VLM annotates the dataset with the probability of the event every 6 frames; in other words, the temporal resolution of our annotation may not be sufficient in this case to build a reliable training set.
Lastly, our model works at $128\,\text{px}$ resolution, which may render barrier hit identification more difficult (e.g., some of the visual evidence for a barrier hit are sparkles coming from the impact, but this may not be visible at low resolution).
A first conclusion that can be drawn from this analysis is that, even if annotating a dataset with a VLM and training an agent for reward prediction is feasible, attention is still required: consistency between the spatial and temporal resolution of the event, of the VLM and of the agent must be guaranteed for successful training. Furthermore, the adoption of a VLM for annotating sparse rewards does not change the sparse nature of the rewards itself, as it is evident from the fact that the distribution of $R_3$ remains unbalanced; moving from sparse to dense rewards, in other words, requires changing the rewards, not simply using a VLM for annotating the dataset.

\begin{figure*}[h!]
    \begin{center}        
    \begin{overpic}[width=0.89\linewidth, trim = 4cm 5cm 3cm 1cm, clip]{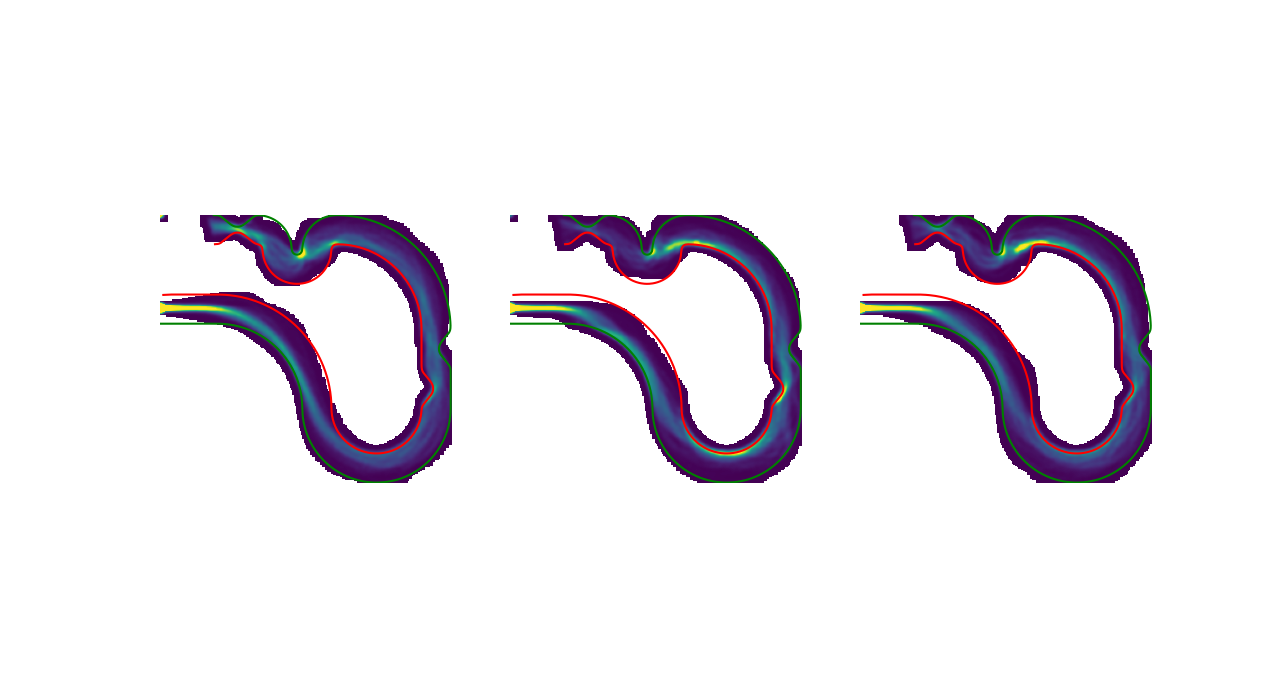}
    \put(0,14){\color{black} \scriptsize Start}
    \put(1,23){\color{black} \scriptsize Finish}    
    \put(8,31){\color{black} \scriptsize Unconditioned}
    \put(40,31){\color{black} \scriptsize Minimize Middle}
    \put(75,31){\color{black} \scriptsize Maximize Middle}
    \end{overpic}\\
    \begin{overpic}[width=0.89\linewidth, trim = 4cm 5cm 3cm 1cm, clip]{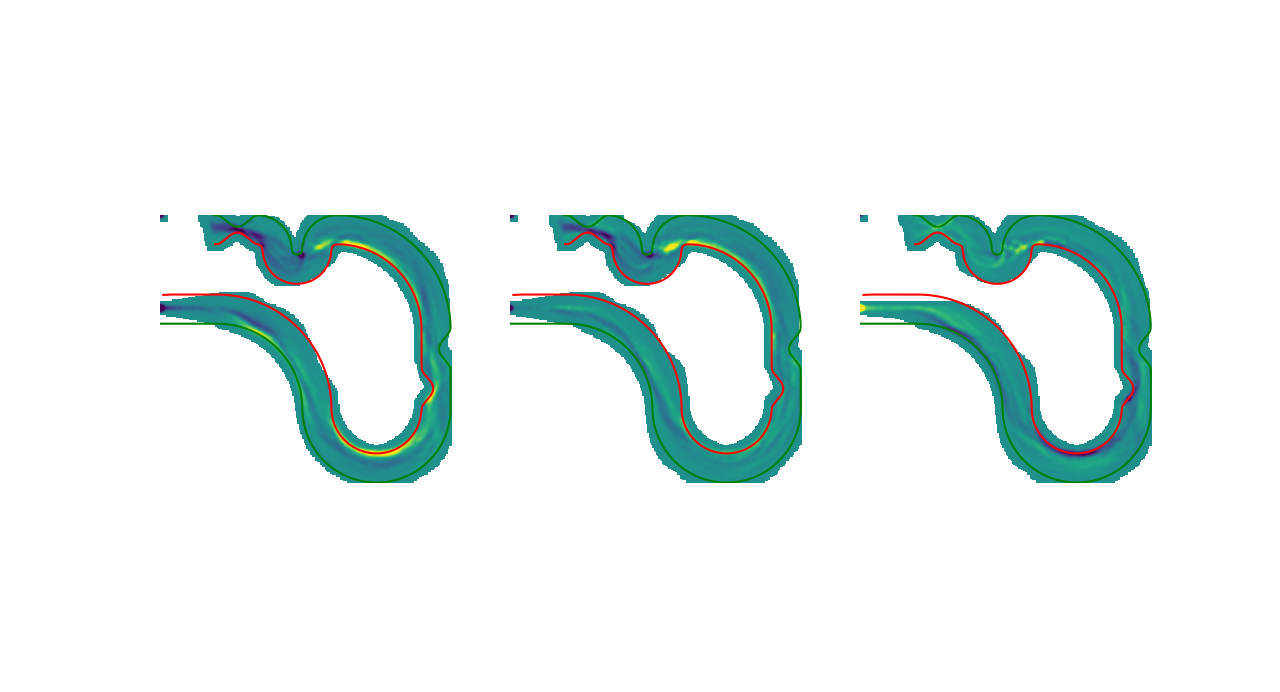}
    \put(8,32){\scriptsize \color{black} \scriptsize Minimize Middle - Unconditioned}
    \put(40,32){\color{black} \scriptsize Maximize Middle - Unconditioned}
    \put(75,32){\color{black} \scriptsize Maximize Middle - Minmize Middle}
    \end{overpic}    
    \caption{The upper panels show the probability of the car position on the test track for the unconditioned model and for the conditioned model forced to minimize or maximize $r_1$. The bottom panels show the differences between the models. The green and red lines identifies the track right and left barriers. Probabilities larger than 0 out of the track limits are due to filtering for visualization purpose.}
    \label{fig:hist}
    \end{center}    
\end{figure*}
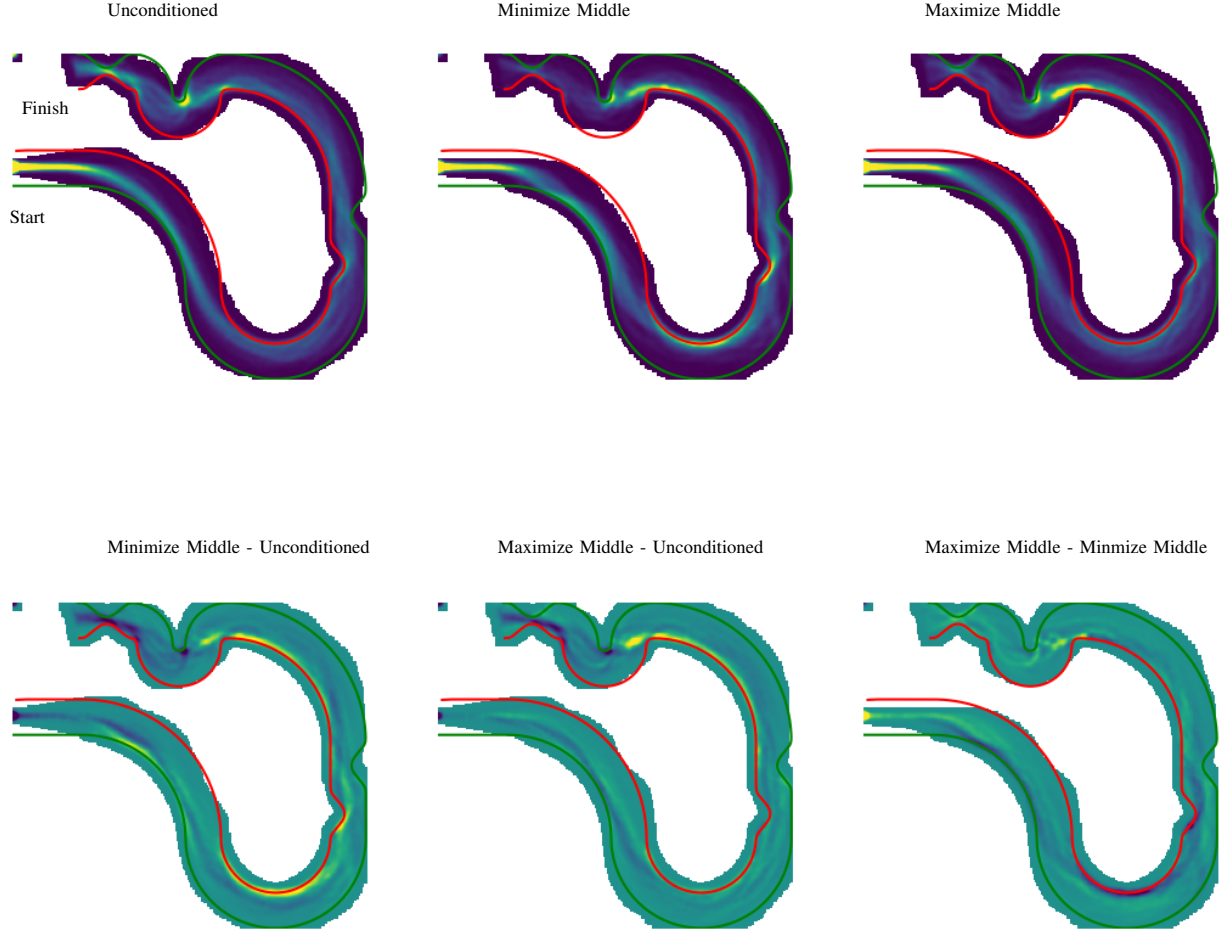

The analysis of Table \ref{tab:metrics} also highlights another interesting fact: the performance of the agent in reward estimation may change with the conditioning signal (the most evident case is a decrease from $26\%$ to $13\%$ in $F1$ for $Q_3$).
This seems to be associated with the interaction between the conditioning signal and the agent behavior. Clearly, as the driving style changes when asking the car to stay in the middle of the road or away from it, the distribution of the rewards changes accordingly and with it the difficulty of estimating them.
The distribution of the training data may play a fundamental role in this: since we do not force driving in the middle of the road (or away from it) during dataset acquisition, one condition may be less represented than the other and consequently generate more or less training data for the conditioned agent with different desired returns.
In literature this problem has been tackled by predicting the distribution of the next desired returns and sampling from it (see~\cite{lee2022multigame}), while here we are providing desired returns without sampling from their expected distribution. 
A larger set of training data and a careful analysis of their distribution are likely needed to guarantee that we do not provide out-of-distribution inputs at inference time and consequently decrease the accuracy of the estimated rewards.

The analysis of the average rewards $R_j$ in Table~\ref{tab:metrics} further suggests what is likely the most important result here, i.e., the fact that the model behavior changes with the conditioning signal, thus confirming that training a conditioned model on VLM annotated data can be successfully performed.
In fact, while the agent conditioned to minimize $r_1$ keeps the vehicle in the middle of the road 37\% of the time, the model conditioned to maximize it stays in the center of the road for 43\% of the time. The probability of hitting the lateral barriers, which is clearly connected with the position of the car, also drops from 11\% to 9\%.
Fig.~\ref{fig:hist} shows the distribution of the car position on the test track for an unconditioned agent (trained with behavioral cloning without using annotated rewards) and for the conditioned agent (trained on returns estimated from the VLM output) asked to minimize or maximize $r_1$.
In the leftmost panel (simple behavioral cloning) the car grabs the curbs in the curves, following the likely optimal racing trajectory.
The second panel from the left panel shows that, when the model is asked to minimize $r_1$, the car indeed stays closer to the edges of the track (see for instance the concentration of the probability close to the red border for the south curve), whereas in the rightmost panel the distribution of the car position is more spread across the track width.
Training conditioned models on VLM-annotated dataset therefore seems feasible.

The three lower panels in the same figure show, on the other hand, a slightly more complicated story: although the difference between the conditioned models and the baseline is clear and aligned to expectations in the first curve of the track, it also tends to vanish in the second part of it.
Our interpretation is that this is due to covariance shift: the model drives according to the provided instruction in the first part of the track where it's easy to stay within the training distribution.
As soon as small errors accumulate, though, the level of confidence of the model decreases and conditioning no longer works as expected.
This issue is related to the simple training algorithm, not to the annotation and training pipeline presented here.
Rather than pointing out that solutions to the distribution shift issue already exist, we want to highlight, though, that conditioned model training requires more data than simple behavioral cloning (especially when some noise on the training conditioning signal is present as in our case).
The positive aspect within the context of our framework is that new data required for training can be automatically generated by a partially trained agent and then annotated by a VLM with no human supervision.
This paradigm has strong similarities with DAGGER~\cite{ross2011reduction}, a classical algorithm where an expert intervention is required to supervise and correct the agent behavior on slightly off-distribution trajectories.
In our case the expert executor providing the optimal trajectory is not needed anymore: an expert VLM annotator (or supervisor) is instead sufficient to generate the training signal.

\section{Conclusion}

We have shown here preliminary evidence that a conditioned policy for playing a video game can be trained on VLM annotated data.
This has several advantages over traditional RL that requires access to the video game engine and often relies on sparse rewards, while our approach uses a set of rewards that are easy-to-understand for humans and may therefore allow the development of playing policies within a natural communication paradigm between the machine and the user.
Although our experiments show encouraging results, several limitations call for more research effort to make our approach practical to use. We have shown, for instance, that some rewards may remain sparse and lead to unbalanced training datasets, and that attention must be paid to the distribution of returns and training data in order to train conditioned agents that do not go out of distribution at inference time.
These limitations are defining the next steps in our research activity.


\bibliographystyle{plain} 
\bibliography{references} 

\end{document}